# A Transfer Learning based Feature-Weak-Relevant Method for Image Clustering

Bo Dong, Xinnian Wang
Dalian Maritime University
Dalian, China
{dukedong,wxn}@dlmu.edu.cn

**Abstract.** Image clustering is to group a set of images into disjoint clusters in a way that images in the same cluster are more similar to each other than to those in other clusters, which is an unsupervised or semi-supervised learning process. It is a crucial and challenging task in machine learning and computer vision. The performances of existing image clustering methods have close relations with features used for clustering, even if unsupervised coding based methods have improved the performances a lot. To reduce the effect of clustering features, we propose a feature-weak-relevant method for image clustering. The proposed method converts an unsupervised clustering process into an alternative iterative process of unsupervised learning and transfer learning. The clustering process firstly starts up from handcrafted features based image clustering to estimate an initial label for every image, and secondly use a proposed sampling strategy to choose images with reliable labels to feed a transfer-learning model to learn representative features that can be used for next round of unsupervised learning. In this manner, image clustering is iteratively optimized. What's more, the handcrafted features are used to boot up the clustering process, and just have a little effect on the final performance; therefore, the proposed method is feature-weak-relevant. Experimental results on six kinds of public available datasets show that the proposed method outperforms state of the art methods and depends less on the employed features at the same time.

## 1 Introduction

Cluster analysis [1] plays an important role in data mining and machine learning, whose mission is to partition objects into independent groups or clusters. Image clustering aims to classify images into categories based on image similarity, and it is valuable for future studies especially in machine learning. Desirable clustering results can ensure efficiency of further explorations. There are many classic clustering methods, such as K-means [18], GMM [19], Mean-shift [45], Spectral Clustering [46], DBSCAN [47], etc. Whatever kinds of clustering methods, the employed clustering features still have great effects on clustering results.

Two types of clustering features are always used for image clustering:

- Handcrafted features

According to prior knowledge of the sampled images, handcrafted features, such as color histogram [48], SIFT [49], HOG [34], Gabor [33], and LBP [50], can be applied for a quick image clustering. However, the clustering results can be unsatisfying when the feature space is indistinguishable for part of clusters. When implemented with different features, image clustering performs unstably because of its strong relevance to the employed features.

- Unsupervised encoding based features

Unsupervised Learning can be used for image encoding/decoding. The encoded codes are usually robust to noise and preserves information almost matches the original data. Therefore, the codes can be employed in image clustering. Unsupervised learning based clustering is mainly classified into two types: encoding based on linear models and encoding based on nonlinear models. The former type tends to map original data to a more distinguishable feature space by linear optimization. These methods can be well performed in data mapping, but linear models may be too simple to handle various image sets for clustering. For the latter method, deep neural networks (DNNs) based unsupervised learning techniques have been widely used for data preprocessing, especially DNNs based auto-encoding techniques. Though these techniques are good at data remapping and dimension reduction, there are no cooperative links between clustering and learning. That may lead to undesirable results at the same time [31]. Therefore, it is considered that mapping and clustering can be implemented jointly. The typical algorithms based on the 'joint idea' include JNKM [30], DEC [27] and DCN [31]. Actually two types of optimization objectives are considered as the minimum reconstruction errors and clustering losses. However, there exists a contradiction in the joint process: the minimum reconstruction errors guarantee well preserved information including interference information of images, which may lead to big clustering losses. At the same time, the joint process consumes too much time on DNN optimization, which limits further improvements in this field.

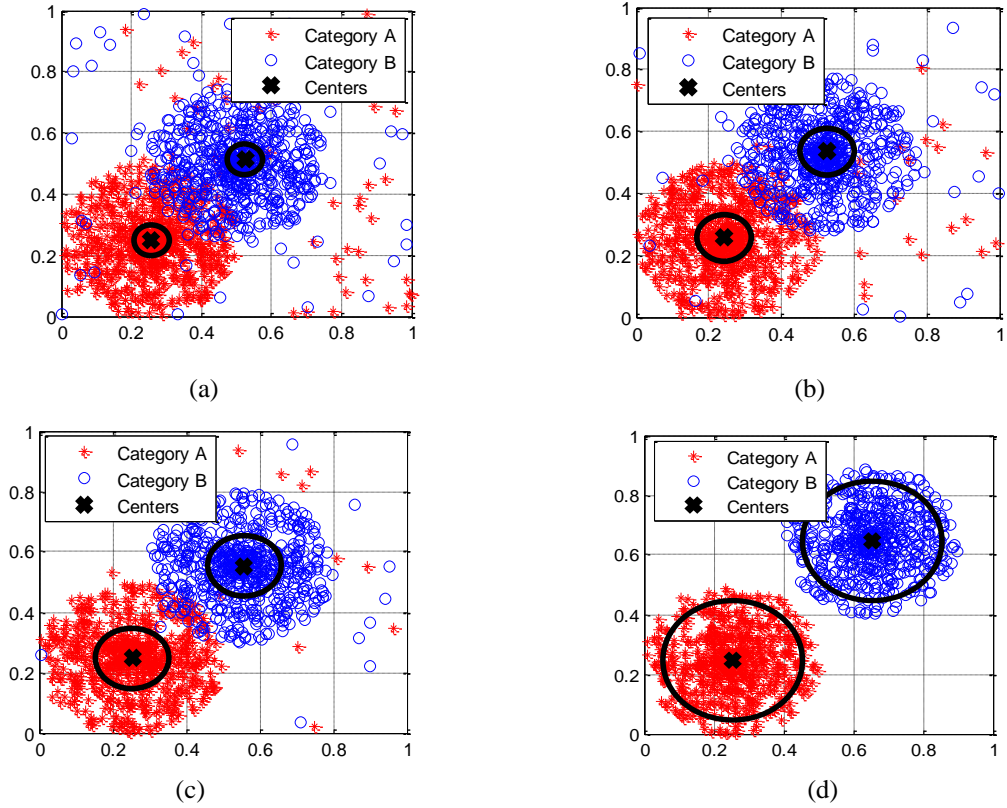

Fig. 1. The illustration for Transfer Learning based Feature-Weak-Relevant clustering method(circled samples are considered to be with correct labels, and the others not): (a) initial feature distribution and the related cluster center in two different clusters; (b) the learned feature distribution after $1^{st}$ iteration; (c) the learned feature distribution after $2^{nd}$ iteration; (d) the learned feature distribution after $3^{rd}$ / final iteration of transfer learning based clustering

Compared with handcrafted features and unsupervised encoding based features, features learned by supervised learning are more representative and discriminative, but the process of acquiring features needs a large amount of labeled images which are not provided for image clustering. An alternative way is to generate clustering features using a Convolutional Neural Networks (CNN) [21] pre-trained with labeled images. However, this idea may not work out well when the labeled and unlabeled images are independently distributed in feature space. Fortunately, fine-tuning the pre-trained CNN [20], as one type of transfer learning methods, is a better way to overcome the shortcomings of direct features generation. At the same time, this process still needs supervised information such as labels.

Our idea is to use images with credible labels obtained by handcrafted features based clustering for booting the transfer learning process, and the features learned by transfer learning can be used for another round of clustering to get more images with credible labels which are used for next round of transfer learning. In this manner, more and more sampled images with credible labels are obtained, and image clustering can be iteratively optimized. An illustration of the idea is shown in Figure 1. To implement the idea, we propose a transfer learning based feature-weak-relevant method for image clustering. The proposed method aims to make image clustering depend less on handcrafted features by transforming an unsupervised learning process into an alternative iterative process of unsupervised learning and transfer learning. Experimental results on six kinds of public available datasets show that the proposed method outperforms state of the art methods and depends less on the employed features at the same time.

## 2  Background and Related Work

Assuming there is a set of images $\mathbf{X} = \{\mathbf{I}_i\}_{i=1,2,\cdots,N}$, where $\mathbf{I}_i \in \mathbf{R}_+^{m \times n}$, the process of image clustering is to participate $\mathbf{X}$ into $K$ ($K \in \mathbf{Z}_+$) disjoint sets $\mathbf{C}$, where $\mathbf{C} = \{\mathbf{C}_k\}_{k=1,\cdots,K}$. At the same time, $\bigcup_{k=1}^{K} \mathbf{C}_k = \mathbf{X}$ and $\mathbf{C}_r \cap \mathbf{C}_q = \varnothing$ when $r \neq q$. For $\mathbf{I}_i \in \mathbf{X}$, we use $L_i \in \{1, 2, \cdots, K\}$ to denote the label of the cluster to which $\mathbf{I}_i$ belongs. In the case of unsupervised image clustering, $L_i$ is unavailable and should be predicted by the image clustering algorithm.

Typically, a clustering model can be described as follows:

$$\min_{\mathbf{M} \in \mathbf{R}^{D \times K}, s_i \in \{0,1\}^K} \sum_{i=1}^{N} d(\mathbf{f}_i, \mathbf{M}\mathbf{s}_i) \tag{1}$$

Here, $\mathbf{f}_i \in \mathbf{R}^D$ denotes descriptive features of $\mathbf{X}_i$. $\mathbf{f}_i = g(\mathbf{I}_i)$, where $g(\cdot)$ denotes a feature extraction function. $\mathbf{M}$ is a descriptive matrix for each cluster, and each column of $\mathbf{M}$ denotes one kind of specified cluster information, such as the centroid. $\mathbf{s}_i$ is a one-hot vector for a single cluster, and $d(\cdot)$ refers to the distance measurement model. Most of cluster algorithms are implemented based on (1).

Unsupervised encoding based $g(\cdot)$ is mostly studied in mapping $\mathbf{I}_i$ towards $d(\cdot)$-friendly spaces. Some of the encoding methods may introduce other constraints for mapping $\mathbf{F} = \{\mathbf{f}_i\}_{i=1,2,\cdots,N}$, which makes $g(\cdot)$ more complex as shown in (2):

$$\mathbf{f}_i = t[g(\mathbf{I}_i)] \tag{2}$$

where $t(\cdot)$ is the mapping function.

Typical linear-based $t(\cdot)$s include Principal Component Analysis (PCA) [9], Canonical Correlation Analysis (CCA) [10], Nonnegative Matrix Factorization (NMF) [11] and Sparse Representation (SR) [12]. Nonlinear-based $t(\cdot)$s include Stacked Auto Encoder (SAE) [13], Sparse Auto Encoder [14], Denoising Auto Encoder (DAE) [15] and Contractive Auto Encoder (CAE) [16]. As shown in (1) and (2), there are no mathematical connections between feature extraction and clustering, which makes the clustering results not satisfying enough and extracted features less representative. However, the mapping techniques are still popular among the researches on image clustering. Recently, instead of proposing the handcrafted dense/sparse feature models, in order to embed effective mathematical connections between features and clustering, researchers [27-29] have been studying on linear/nonlinear generative model from the latent space to the data domain while clustering.

One of the most impressive achievements is shown in [30]. Yang et al. (2017) define (1) as follows:

$$\min_{\mathbf{M} \in \mathbf{R}^{D \times K}, s_i \in \{0,1\}^K, \mathbf{W}, \mathbf{F}} \lambda \sum_{i=1}^{N} d(\mathbf{f}_i, \mathbf{M}\mathbf{s}_i) + \|\mathbf{X} - \mathbf{W}\mathbf{F}\| + r_1(\mathbf{F}) + r_2(\mathbf{W}) \tag{3}$$

where $\lambda (\lambda \in \mathbf{R}_+)$ is a parameter for balancing data fidelity and the latent cluster structure. $r_1(\cdot)$ and $r_2(\cdot)$ are pre-defined regularization parameters. In this way, $g(\cdot)$ and $t(\cdot)$ can be simplified into a trainable and question-related linear model.

For further study, Yang et al.(2017) believe the data generating process is too complex to be precisely approximated based on a linear transformation, and the mapped features may not be representative enough. Therefore, in [31], they introduce Deep Neural Networks (DNN) into (3) as follows:

$$\min_{\mathbf{M} \in \mathbf{R}^{D \times K}, s_i \in \{0,1\}^K} \sum_{i=1}^{N} (\lambda d(g(\mathbf{I}_i), \mathbf{M}\mathbf{s}_i) + \ell(p(g(\mathbf{I}_i)), \mathbf{I}_i)) \tag{4}$$

$\ell(\cdot)$ is the least-squared loss function for measuring the reconstruction error. $g(\cdot)$ and $p(\cdot)$ represent the encoding part and the decoding part respectively. It is an excellent idea to perform clustering with dimensionality reduction based on nonlinear SAE jointly for handcrafted features are no longer needed. SAE based (4) can automatically map from high-dimensional data to its latent space and the loss function of K-means clustering guarantees that the mapped feature is suitable for effective clustering.

However, for all encoding based $g(\cdot)$s or $t(\cdot)$s, the encoded $F$ are redundant [17] because each $I_i$ consists of clustering relevant data $Ig_i$ and irrelevant data $Ib_i$ as shown in (5):

$$I_i = Ig_i + Ib_i \tag{5}$$

Hence, there exists a contradiction in (4): the minimum reconstruction error can guarantee $F$ maintain all image information including interference features have negative effects on image clustering. It is an efficient way to embed $g(\cdot)$ into clustering process as mentioned in the latest researches [27~31], and there exists improvements in feature simplification and clustering. In addition, SAE based clustering is working with high time and memory consumption in DNN training.

## 3 Transfer Learning based Feature-weak-relevant Image Clustering

### 3.1 Formulations of the prototype

We hope images in the same cluster to be more similar to each other than to those in other clusters, and the cost function can be described as (6):

$$\min_{\mathbf{M} \in R^{D \times K}, s_i \in \{0,1\}^K, g(\cdot)} \sum_{i=1}^{N} d(g(\mathbf{I}_i), \mathbf{M}s_i) + \\ \alpha \sum_{i=1}^{N} \sum_{j=1}^{N} d_{in}(g(\mathbf{I}_i), \{g(\mathbf{I}_j)\}_{L_j = L_i}) - \\ \beta \sum_{i=1}^{N} \sum_{j=1}^{N} d_{out}(g(\mathbf{I}_i), \{g(\mathbf{I}_j)\}_{L_i \neq L_j}) \tag{6}$$

where $d_{in}(\cdot)$ represents the model for measuring image distance in the same clusters and $d_{out}(\cdot)$ represents the model for measuring image similarity in different clusters. $\alpha (\alpha \in \mathbf{R}_+)$ and $\beta (\beta \in \mathbf{R}_+)$ are weight parameters. $g_t(\cdot)$ is an ideal trainable function for generating ideal features. However, (6) is unsolvable for image clustering because image clustering and representative feature learning are cause-and-effect related.

- To gain clustering results with high accuracy, $g(\cdot)$ should be the same as $g_t(\cdot)$, which is built or trained with $d_{in}(\cdot)$ and $d_{out}(\cdot)$ at first;
- To train $g(\cdot)$ well, large amounts of labeled images are required at first;
- To predict effective cluster labels, $g(\cdot)$ must be defined at first.

However, equation (6) can be solved in an iterative form by assuming an initial feature extraction function $g_0(\cdot)$, and the iterative form of equation (6) is:

$$\min_{\mathbf{M} \in R^{D \times K}, s_i \in \{0,1\}^K, g_t(\cdot)} \sum_{i=1}^{N} d(g_{t-1}(\mathbf{I}_i), \mathbf{M}^t s_i^t) + \\ \alpha \sum_{i \in v_t} \sum_{j \in v_t} d_{in}(g_t(\mathbf{I}_i), \{g_t(\mathbf{I}_j)\}_{L_j^t = L_i^t}) - \\ \beta \sum_{i \in v_t} \sum_{j \in v_t} d_{out}(g_t(\mathbf{I}_i), \{g_t(\mathbf{I}_j)\}_{L_i^t \neq L_j^t}) \tag{7}$$

where $t \in \mathbb{Z}^+$ denotes the iteration counter. $v_t$ denotes the image sample set that have reliable labels at the $t$th round of iteration, and $L_i^t, L_j^t \in \{1, 2, \cdots, K\}$ denote labels of the sample images belonging to $v_t$. $g_t(\cdot)$ is the feature function to be learned.

The alternative iteration solution of equation (7) is as follows:

- $g_0(\cdot)$ is initialized to be a handcrafted feature extracting function, such as SIFT extractor, HOG extractor.
- $M^t$ and $s^t = \{s_1^t, \cdots, s_N^t\}$ are obtained by minimizing the first term of equation (7), i.e.:

$$M^t, s^t = \min_{M \in R^{D \times K}, s_i \in \{0,1\}^K} \sum_{i=1}^{N} d\left(g_{t-1}(I_i), M^t s_i^t\right) \quad (8)$$

Assuming $L^* = \{L_i^*\}_{i=1,2,\cdots,N}$ is the true labels of $X$, and $\hat{L}^* = \{\hat{L}_i^*\}_{i=1,2,\cdots,N}$ is the estimated labels of $X$, we can get $\hat{L}^*$ the clustering optimization.

- $v_t$ is determined by a sampling function $\varphi(\cdot)$ whose goal is to choose reliable labelled samples with $M^t$ and $s^t$ as input, i.e.:

$$v = \varphi(M^t, s^t) \quad (9)$$

Once $v_t$ is determined, $L^t$ is determined. We define $\varphi(\cdot)$ mathematically to make sure part of the labels $L_i^* = \hat{L}_i^*$ or acceptably $L_i^* \approx \hat{L}_i^*$ to solve (8). Here, $v(v \in \{0,1\}^N)$. If $v_i = 1$, we ensure $\hat{L}_i^* \approx L_i^*$ acceptable. If $v_i = 0$, we believe that the predicted $\hat{L}_i^*$ is not reliable.

- $g_t(\cdot)$ is learned by minimizing the 2nd and the third term of equation (7), which is to say:

$$g_t(\cdot) = \min_{g_t(\cdot)} (\alpha \sum_{i \in v_t} \sum_{j \in v_t} d_{in}(g_t(I_i), \{g(I_j)\}_{L_j^t = L_i^t}) - \beta \sum_{i \in v_t} \sum_{j \in v_t} d_{out}(g_t(I_i), \{g(I_j)\}_{L_j^t \neq L_i^t})) \quad (10)$$

As long as labelled images are provided, equation (10) can be solved based on CNN classification models and this process is shown as follows:

$$\mathbf{Nets} = train(\mathbf{X}_t, v\hat{L}^*) \quad (11)$$

$\mathbf{X}_t (\mathbf{X}_t \in \mathbf{X})$ denotes the reliable labelled samples at the $t$th round of iteration. Features that are more valid can be generated by computing values from any layers of the fine-tuned net, which can be described as follows:

$$F_t = g_t(\mathbf{X}) = getblob(\mathbf{X}, \mathbf{Nets}) \quad t > 0 \quad (12)$$

Alternatively, we can just use the trained model for updating $\hat{L}^*$ instead of (8) by prediction as (13) shows:

$$\hat{L}^* = predict(\mathbf{X}, \mathbf{Nets}) \quad (13)$$

As features are generated by $\mathbf{Nets}$ based on (12), the updated $\hat{L}^*$ will be more reliable. We can train models to be more effective using more images with credible $\hat{L}^*$. However, it may lead to an over fitted CNN model when only a small amount of images is supplied for training. Commonly, we cannot ensure the credible labelled images are enough or not. Thus, there exists a risk generating unsatisfying models when solve (9) directly. In order to avoid the problems of divergence and overfitting caused by distribution of training data, the transfer learning method of fine-tuning a pre-trained classification model is adopted. And pre-trained structures include LeNet [22], AlexNet [23], VGG [24], GoogleNet [25], ResNet[26] and others.

Among the above existing structures, AlexNet is one of the most popular structures. Though AlexNet only consists of 5 convolutional layers and 3 fully connected layers, it performs well in the ILSVRC-2012 with a top-5 test error rate of 15.3%. Assuming the pre-trained AlexNet is $\mathbf{ANets}$, model (11) can be then changed as follows:

$$\mathbf{Nets} = finetune(\mathbf{X}_t, v\hat{L}, \mathbf{ANets}) \quad (14)$$

In conclusion, we need to initialize $\mathbf{X}$ first for booting transfer learning before the feature optimization and then implement the optimization by fine-tuning models and increasing the number of images that satisfy $\hat{L}_i^* \approx L_i^*$. Therefore, we define the initialization part as **the boot stage**, and the initial features as the boot features. All other steps are called **the main stage**, and the final features can be more question-relevant. At last, model (6) can be solved by iteratively updating $\hat{L}^*$ based on transfer learning.

### 3.2 $\varphi(\cdot)$ Related Definitions

According to the definition of $\varphi(\cdot)$, $v_i$ can be described as follows:

$$v_i = c(f_i, \{f_j\}_{\hat{L}_j^{t*}=\hat{L}_i^{t*}}) \tag{15}$$

where $c(\cdot)$ is a function for measuring cohesion between $I_i$ and $I_j$ with the same label $\hat{L}_i^{t*}$. As $v_i$ is a binary label for grouping $\hat{L}$, $c(\cdot)$ can use the thresholding method for binarization as follows:

$$c(f_i, \{f_j\}_{\hat{L}_j^{t*}=\hat{L}_i^{t*}}) = \begin{cases} 1 & c_b(f_i, \{f_j\}_{\hat{L}_j^{t*}=\hat{L}_i^{t*}}) < \Delta \\ 0 & else \end{cases} \tag{16}$$

where $c_b(\cdot)$ represents inner distance evaluation models for images with same labels. According to clustering theories, there is a probability that images with features that densely distributed around the same center belong to the same class. The closer images are to the same center, the greater the above probability will be. The closer images $X$ with $v = \boldsymbol{1}^{N_1}$ are called Cluster Center Neighboring sets (CCNs) and the rest with $v = \boldsymbol{0}^{N_2}$ are called Cluster Center Distant sets (CCDs). $N_1, N_2 \in \mathbf{Z}_+$, and $N_1 + N_2 = N$. A simple $c_b(\cdot)$ can be defined as (17):

$$c_b(f_i, \{f_j\}_{\hat{L}_j^{t*}=\hat{L}_i^{t*}}) = \|f_i - Ms_i\| \tag{17}$$

There are many alternatives for distance measuring in (17), such as Chess Board Distance, City Block Distance, etc.

In (16), the threshold $\Delta$ is important for binarization. A proper threshold should be given based on the clustering features before optimization, and it varies with different types of handcrafted feature. Therefore, a constant $\Delta$ can be initialized easily, but it is not effective for all features. To avoid the issue, we define $\Delta$ as a relative ratio that can be also initialized simply. Instead of referring to the results of $c_b(\cdot)$ directly, we sort the results on same labels in descending order as (18) shows.

$$\boldsymbol{\rho}_{\hat{L}_i^{t*}} = sort(\{c_b\}_{\hat{L}^{t*}=\hat{L}_i^{t*}}) \tag{18}$$

where $\boldsymbol{\rho}_{\hat{L}_i^{t*}}$ is a vector containing the sorted distances. We set $c_b(f_i, \{f_j\}_{\hat{L}_j^{t*}=\hat{L}_i^{t*}})=1$ if $c_b(f_i, \{f_j\}_{\hat{L}_j^{t*}=\hat{L}_i^{t*}})=1$ is in the top $\Delta\%$ of $\boldsymbol{\rho}_{\hat{L}_i^{t*}}$. In this way, $\Delta$ is considered as a belief ratio indicating there are $\Delta\%$ CCNs in every cluster. The relative threshold is more robust to variation of the employed handcrafted clustering features.

### 3.3 Optimization Procedure in Main Stage

In section 3.1, we convert (6) into an iterative optimization model that can be solved by iteratively updating $\hat{L}^{t*}$. In order to make the optimization implementable, we need to add following three constraints (assuming $k$ is the iteration counter, $k \in \mathbf{Z}_+$):

- CCNs and CCDs can never be the same until the optimization is done, and $\|v_k \cap v_{k+1}\|_1 \geq 0$.
- The number of images in CCNs must be increasing during the optimization, and $\|v_k\|_1 < \|v_{k+1}\|_1$.
- The CCNs cannot be modified after confirmation, and $\|v_k \cup v_{k+1}\|_1 = \|v_{k+1}\|_1$.

### 3.4 Algorithm Description

Based on constrains in section 3.3 and the implementation analysis, we can update $\hat{L}^{t*}$ as Algorithm 1 shows.

$k_{max}$ represents the maximum iterative count. $\delta(\delta \in [0,1])$ is an adjusting acceleration factor. $\mathbf{X}_t$ can be forcedly enlarged by gradually relaxing $\Delta$ when $\delta > 0$ ($\Delta + \delta \leq 1$). So $\delta$ the method can be considered as an optional factor if fast results are required. Although the method of gradually relaxing is regarded as a local optimized result proved from the experimental results, it still outperforms than the stage of the arts methods.

---

**Algorithm 1 Iterative Updating of $\hat{L}^{t^*}$**

**Input:** $\mathbf{X}$, $k_{max}$, $\Delta$, $\delta$, **ANets**

**Output:** $\hat{L}^*$

**Steps:**

S1: Initialize handcrafted features $F$;

S2: Initialize $k=0$, $v_0=\mathbf{0}$, $\mathbf{X}_t=\varnothing$ and $\mathbf{X}_s=\mathbf{X}$;

S3: Compute $v_k$ of $\mathbf{X}_s$ by (15);

S4: Split $\mathbf{X}_s$ into CCN sets $\mathbf{X}_{st}$ and CCD sets $\mathbf{X}_{ss}$ with $v_k$;

S5: Update $\mathbf{X}_t=\mathbf{X}_t\bigcup\mathbf{X}_{st}$, $\mathbf{X}_s=\mathbf{X}_{ss}$;

S6: Fine-tune **Nets** with $\mathbf{X}_t$ by (14);

S7: Update $\hat{L}^{t^*}$ of $\mathbf{X}_s$ by (13);

S8: Update features $F_s$ of $\mathbf{X}_s$ by (12);

S9: $\Delta = \Delta + \delta$;

S10: $k = k+1$;

S11: Repeat S3 ~S10 until $\mathbf{X}_s$ is $\varnothing$ or $k > k_{max}$.

---

# 4 Experiments

In this section, various real-world data sets, different types of initial feature sets and clustering methods are applied to the proposed method for effectiveness validation, respectively. In sections 4.1~4.3, we provide detailed information of implementing relevant experiments, including baseline methods, evaluation models and public image sets for clustering. In section 4.4 different comparative experimental results are respectively analyzed. Then we investigate the sensitiveness of the parameter Δ and δ in section 4.5. Meanwhile the experiments on time consumption are analyzed in section 4.6. All fine-tuning experiments are implemented with Caffe [32] and the detailed information of the prototxt file related to fine-tuning is available at https://github.com/DukeDong/Transfer-Learning-Based-Image-Clustering-.git.

## 4.1 Baseline Methods

Clustering features and methods have great effect on the performance of an image clustering algorithm, so we choose the following features, methods and their combinations for evaluations.

Table 1. Combinations of clustering features and clustering methods

| ID | Clustering Features | Clustering Method |
|---|---|---|
| #1 | RP | K-means |
| #2 | GF | |
| #3 | HOG | |
| #4 | RP+PCA | |
| #5 | RP | GMM |
| #6 | GF | |
| #7 | HOG | |
| #8 | RP+PCA | |
| #9 | DCN (RP+K-means) | |

**Clustering features:**

- **Raw Pixels (RP)**: pixel values without processing.
- **Gabor Features (GF)**: features generated by Gabor transform [33]. This kind of feature shows the information in spatial and frequency domain.
- **HOG Features: histogram of gradient** [34]. HOG represents edges of objects in images, and it is robust to geometrical and optical deformation.
- **RP + PCA**:  feature vectors generated by PCA. The combination is still applied in face recognition [35].

**Clustering methods:**

- **K-means** [18]: the classic cluster method.
- **Gaussian Mixture Model (GMM)** [19]: the cluster method based on a latent variable model.
- **Deep Clustering Network (DCN)** [31]: a new method that can jointly optimize dimension reduction and clustering. It is accomplished via learning a deep neural network to approximate any nonlinear function.

There are nine kinds of combinations between different features and methods for image clustering including one method based on DCN, which are shown in table 1.

### 4.2 Evaluation Models

There are mainly two kinds of models for performance evaluation of a clustering method, which are external index and internal index. The applied models with external index include Jaccard Coefficient (JC) [36], Fowlkes and Mallows Index (FMI) [37], Normalized Mutual Information (NMI), Clustering Accuracy (ACC) [38] and Adjusted Rand Index (ARI) [39]. As to internal index, we employ Davies-Bouldin Index (DBI) [40]. Comparing DBI directly is meaningless when different features are employed for clustering. Therefore, we calculate final DBI by normalizing DBI gained in the boot stage. In this way, the normalized DBI implies a reduction of the initial DBI.

Besides, one of our goals is to weaken the effect of initial features. Another evaluation factor is required to examine how much effect the proposed method can weaken the initial features. Assuming clustering results with different initial features for a clustering method are available, the evaluation factor $\tau$ can be defined as (19):

$$\tau_i = \frac{\sigma(\boldsymbol{\eta}_i)}{E(\boldsymbol{\eta}_i)} \tag{19}$$

where $\tau_i$ is the index difference rate, $\boldsymbol{\eta}_i \in \mathbf{R}^9$ is a vector containing the $i$ th external index values computed from the results of different features, and $i \in \{1,2,3,4,5\}$ represents the ID of applied evaluation indexes respectively named JC, FMI, NMI, ARI, and ACC. $\sigma(\cdot)$ and $E(\cdot)$ compute standard deviation and mean values of $\boldsymbol{\eta}_i$. Small value of $\tau_i$ means that the corresponding method does well in weakening the effect of initial features.

### 4.3 Data Sets

Image clustering methods need to handle many different kinds of objects. In order to evaluate the generalization of the proposed method, experiments are conducted on six typical public image sets:

- MNIST [22]: MNIST dataset has 70,000 data samples of handwritten digit ranging from 0 to 9.
- Caltech Categories: Caltech Categories offers 6 image classes, including 652 images for Cars, 826 images for Motorcycles, 1,074 images for Airplanes, 450 images for unique faces of around 27 people, 186 images for 3 species of Leaves and 550 images for assorted Scenes.
- Caltech 101 [41]: Caltech 101 contains 101 categories including certain species of animals, planes, chairs, medical images and so on.
- Stanford Dogs Dataset [42]: Stanford Dogs Dataset contains images of 120 breeds of dogs from around the world.

- FERET [43]: The database contains 1,564 sets of images and 14,126 images in total including image sets for 1,199 individuals and 365 duplicate sets.
- CASIA-WebFace [44]: CASIA-WebFace has 10,575 subjects with 49,414 images in total.

## 4.4 Performance Evaluations on Six Public Available Datasets

### 4.4.1 Performance on Caltech Categories

Table 2. Clustering evaluation on Caltech Categories

| Method | JC | FMI | NMI | ARI | ACC | DBI |
|---|---|---|---|---|---|---|
| #2 | 0.39 | 0.56 | 0.52 | 0.44 | 0.62 | 1.0 |
| #2+Our Method | **0.65** | **0.79** | **0.80** | **0.73** | **0.71** | **0.77** |
| #3 | 0.40 | 0.58 | 0.59 | 0.48 | 0.63 | 1.0 |
| #3+Our Method | **0.61** | **0.75** | **0.79** | **0.69** | **0.75** | **0.72** |
| #4 | 0.19 | 0.32 | 0.24 | 0.15 | 0.43 | 1.0 |
| #4+Our Method | **0.52** | **0.68** | **0.71** | **0.61** | **0.69** | **0.89** |
| #7 | 0.34 | 0.50 | 0.31 | 0.37 | 0.50 | 1 |
| #7+Our Method | **0.64** | **0.78** | **0.80** | **0.72** | **0.70** | **0.39** |

Table 3. Evaluation of $\tau$ on Caltech Categories

| Method | JC | FMI | NMI | ARI | ACC |
|---|---|---|---|---|---|
| #2~#4 | 0.36 | 0.30 | 0.41 | 0.50 | 0.20 |
| Our Method | **0.11** | **0.08** | **0.06** | **0.09** | **0.04** |

Images in Caltech Categories are more complicated than other applied data sets, and different image classes require different features.

Though Caltech Categories only contains six categories, it requires image preprocessing to achieve desirable clustering results. Image preprocessing is essential for feature selection in prior researches. However, in order to make the process more challengeable, we employ #2~#4 and #7 directly without any preprocessing related analysis. Clustering with the feature RP directly is meaningless, so #1 and #5 are omitted. In fact, it is effective enough that GMM performs image clustering with only one type of handcrafted features for comparison. For GMM, we only list out the results of #7. In the experiments, $\Delta$ is set to 0.2, and $\delta$ is set to 0.1, which is applied in sections 4.4.2~4.4.6. In addition, the cluster number is set to six for K-means and GMM.

Table 2 shows the five evaluation indexes. Table 3 shows the index difference rates with different features. The results show that the proposed method performs much better than other methods and meanwhile it weakens the effects of initial features.

### 4.4.2 Performance on Caltech 101

Compared with Caltech Categories, Caltech 101 dataset contains more various categories, but each category has fewer samples, sample number of which is round ninety on average. Therefore, Caltech 101 can be used to validate the robustness to the variation of image numbers and cluster numbers.

We randomly choose 20 categories from Caltech 101 in this part, and repeat the random process for more than 3 times to guarantee the effectiveness of our experiments. Accordingly, the cluster number is set to 20 for K-means and GMM. Here, the category number of 20 is larger than common maximum cluster number for clustering evaluation of 2, 5, 10, etc. More clusters mean more difficult clustering mission. However, when we conduct image clustering with bigger clustering numbers, we suffer from high time consumptions of K-means and GMM. Therefore, cluster number of 20 is a rough but proper for clustering performance evaluation.

Table 4. Clustering evaluation on Caltech 101

| Method | JC | FMI | NMI | ARI | ACC | DBI |
|---|---|---|---|---|---|---|
| #2 | 0.09±0.0060 | 0.17±0.0090 | 0.22±0.020 | 0.08±0.013 | 0.26±0.018 | 1 |
| #2+Our Method | **0.22**±0.0090 | **0.38**±0.024 | **0.47**±0.012 | **0.26**±0.088 | **0.45**±0.092 | 1.87±0.27 |
| #3 | 0.17±0.019 | 0.30±0.028 | 0.43±0.018 | 0.24±0.030 | 0.42±0.033 | 1 |
| #3+Our Method | **0.32**±0.12 | **0.50**±0.016 | **0.69**±0.011 | **0.45**±0.021 | **0.60**±0.026 | 1.09±0.090 |
| #4 | 0.10±0.057 | 0.18±0.090 | 0.29±0.014 | 0.12±0.095 | 0.29±0.011 | 1 |
| #4+Our Method | **0.29**±0.080 | **0.44**±0.098 | **0.61**±0.082 | **0.40**±0.010 | **0.56**±0.090 | 1.35±0.36 |
| #7 | 0.10±0.0053 | 0.19±0.0043 | 0.15±0.071 | 0.10±0.0062 | 0.26±0.021 | 1 |
| #7+Our Method | **0.16**±0.0082 | **0.35**±0.012 | **0.31**±0.022 | **0.14**±0.0010 | **0.37**±0.022 | **0.77**±0.032 |

Table 5. Evaluation of $\tau$ on Caltech 101

| Method | JC | FMI | NMI | ARI | ACC |
|---|---|---|---|---|---|
| #2~#4 | 0.36 | 0.33 | 0.34 | 0.57 | 0.26 |
| Our Method | **0.35** | **0.14** | **0.19** | **0.27** | **0.14** |

\* $\tau$ is computed with all generated indexes of all random sets.

Table 4 shows the detailed evaluation indexes. Table 5 shows the index difference rates with different features.

As the experimental results show, the proposed method not only surpasses other methods, but also with the better external index values. At the same time, the proposed method weakens the effect of initial features. However, the results of DBI are unsatisfying, which means the learned features is not densely distributed in feature space and further studies may improve DBI and clustering results.

### 4.4.3 Performance on Stanford Dogs Dataset

Table 6. Clustering evaluation on Stanford Dogs Dataset

| Method | JC | FMI | NMI | ARI | ACC | DBI |
|---|---|---|---|---|---|---|
| #2 | 0.04±6.0e-4 | 0.07±0.0013 | 0.02±0.0030 | 0.01±8.9e-4 | 0.09±0.0023 | 1 |
| #2+Our Method | **0.08**±0.014 | **0.17**±0.021 | **0.27**±0.024 | **0.12**±0.031 | **0.30**±0.014 | 1.99±0.42 |
| #3 | 0.03±3.9e-4 | 0.06±7.9e-4 | 0.03±0.0022 | 0.01±9.4e-4 | 0.11±0.0029 | 1 |
| #3+Our Method | **0.11**±0.014 | **0.21**±0.021 | **0.31**±0.038 | **0.14**±0.032 | **0.32**±0.028 | 1.32±0.29 |
| #4 | 0.03±1.6e-4 | 0.07±3.2e-4 | 0.04±0.0021 | 0.01±3.0e-4 | 0.11±0.0014 | 1 |
| #4+Our Method | **0.10**±0.0030 | **0.19**±0.0083 | **0.26**±0.019 | **0.12**±0.013 | **0.27**±0.022 | 1.83±0.20 |
| #7 | 0.04±8.1e-4 | 0.08±0.0022 | 0.016±9.7e-4 | 0.01±2.2e-4 | 0.09±0.0021 | 1 |
| #7+Our Method | **0.06**±0.0021 | **0.16**±0.013 | **0.16**±0.031 | **0.05**±0.011 | **0.25**±0.011 | 1.06±0.31 |

Table 7. Evaluation of $\tau$ on Stanford Dogs Dataset

| Method | JC | FMI | NMI | ARI | ACC |
|---|---|---|---|---|---|
| #2~#4 | 0.17 | 0.11 | 0.33 | 0.36 | 0.11 |
| Our Method | **0.16** | **0.10** | **0.09** | **0.09** | **0.08** |

\* $\tau$ is computed with all generated indexes of all random sets.

The proposed method performs well on nature images. Maybe it profits a lot from AlexNet, which is trained with a large number of nature images, and some of the clustered classes have been classified. Thus, a more complicated image set should be applied to make sure that the transfer-learned features do not depend on AlexNet totally. Stanford Dogs, as one of the categories, belongs to the whole datasets for training AlexNet. Clustering on dog images is a process that split one of the trained categories into different clusters. If the results turn out to be satisfying, the corresponding transfer-learned features are believed to be desirable and independent from the original AlexNet.

Stanford Dogs dataset has been built using images and annotations from ImageNet for fine-grained image categorization. The annotations include class labels and bounding boxes. It helps a lot when an image clustering algorithm is provided with bounding boxes, but the bounding boxes are hardly available during unsupervised image clustering. In order to make the experiments more reliable, the box information is omitted in this section. Compared with the Caltech datasets, Stanford Dogs Dataset contains categories representing clusters in a more detailed way, and the average number of sample images in each breed is approximately 172. We randomly choose 20 breeds from the database this time, and repeat the random process for more than 3 times to ensure the effectiveness of the experiments. The cluster number is set to 20 for K-means and GMM.

Table 6 shows the detailed clustering evaluation indexes. Table 7 shows the index difference rates with different features. Similar results to those obtained in section 4.4.2, the results of DBI evaluated with the proposed method are turned out to be even larger. As the learned features can be further explored, the clustering performance is highly possible to be improved. However, the detailed external indexes still imply that the proposed method highly improves the clustering results and weakens the effects of initial features.

### 4.4.4   Performance on FERET

The FERET database is used for face recognition system evaluation. As the collection is progressed in a standard way, the chosen images contain just a little background information and many various face photos with different postures.

Different from the previous datasets mentioned above, FERET is classified individually, which means FERET clustering is a process of unsupervised face recognition. Face image clustering is much closer to the leaf nodes than clustering methods based on Caltech categories and dog breeds. It is a traditional way to build up suitable features when the images can be clearly noticed. In order to make the experiments more reliable, the box information is omitted in this section as well. In this part, we randomly choose 20 individuals from FERET, and repeat the random process for more than 3 times to make sure the effectiveness of the experiments. The cluster number F is directly set to 20 for K-means and GMM.

Table 8. Clustering evaluation on FERET

| Method | JC | FMI | NMI | ARI | ACC | DBI |
|---|---|---|---|---|---|---|
| #2 | 0.04±6.0e-4 | 0.08±0.0020 | 0.04±0.0041 | 0.02±0.0014 | 0.26±0.014 | 1 |
| #2+Our Method | **0.12±0.0085** | **0.26±0.014** | **0.38±0.028** | **0.30±0.039** | **0.53±0.053** | 1.77±0.080 |
| #3 | 0.05±0.0054 | 0.10±0.0095 | 0.11±0.020 | 0.06±0.011 | 0.28±0.0082 | 1 |
| #3+Our Method | **0.17±0.023** | **0.33±0.032** | **0.47±0.022** | **0.37±0.023** | **0.54±0.045** | 1.35±0.11 |
| #4 | 0.15±0.013 | 0.26±0.020 | 0.36±0.019 | 0.22±0.019 | 0.42±0.015 | 1 |
| #4+Our Method | **0.17±0.025** | **0.32±0.034** | **0.43±0.088** | **0.31±0.062** | **0.52±0.047** | 2.24±0.045 |
| #7 | 0.12±0.0045 | 0.22±0.0093 | 0.25±0.033 | 0.18±0.0051 | 0.45±0.020 | 1 |
| #7+Our Method | **0.16±0.011** | **0.30±0.025** | **0.44±0.075** | **0.32±0.033** | **0.54±0.031** | **1.53±0.78** |

Table 9.   Evaluation of $\tau$   on FERET

| Method | JC | FMI | NMI | ARI | ACC |
|---|---|---|---|---|---|
| #2~#4 | 0.76 | 0.67 | 0.99 | 1.1 | 0.27 |
| Our Method | **0.19** | **0.12** | **0.11** | **0.11** | **0.02** |

\*   $\tau$  is computed with all generated indexes of all random sets.

Table 8 shows the detailed clustering evaluation indexes. Table 9 shows the index difference rates with different features. Similar results to those reported from Sections 4.4.3 and 4.4.4, the results of the proposed method in DBI are larger than the corresponding ones obtained from initial methods. All the learned features are with the same dimension of 4096, which is fixed and probably redundant. Meanwhile, the data shows the proposed method performs better and weakens the effect of initial features.

### 4.4.5    Performance on CASIA-WebFace

Images in CASIA-WebFace are collected from the Internet, which is different from image collections in FERET, therefore the CASIA-WebFace images normally carry much more background information than those in FERET. That means it is more challenging for image clustering with CASIA-WebFace.

We randomly choose 20 individuals from CASIA-WebFace in the same way, and repeat the random process for more than 3 times. The cluster number is directly set to 20 for K-means and GMM.

Table 10. Clustering evaluation on CASIA

| Method | JC | FMI | NMI | ARI | ACC | DBI |
|---|---|---|---|---|---|---|
| #2 | 0.08±0.0030 | 0.15±0.009 | 0.09±7.1e-4 | 0.04±0.0017 | 0.20±0.110 | 1 |
| #2+Our Method | **0.14**±0.0090 | **0.25**±0.012 | **0.23**±0.012 | **0.11**±0.028 | **0.30**±0.067 | 1.31±0.53 |
| #3 | 0.11±0.0042 | 0.21±0.011 | 0.23±0.013 | 0.14±0.025 | 0.26±0.083 | 1 |
| #3+Our Method | **0.13**±0.0073 | **0.25**±0.0080 | **0.30**±0.021 | **0.17**±0.0030 | **0.30**±0.043 | **0.80**±0.32 |
| #4 | 0.07±0.0012 | 0.14±0.0013 | 0.10±0.0040 | 0.06±0.035 | 0.17±0.0031 | 1 |
| #4+Our Method | **0.15**±0.0062 | **0.26**±0.0098 | **0.22**±0.0090 | **0.08**±0.010 | **0.24**±0.089 | 1.65±0.41 |
| #7 | 0.09±0.0041 | 0.17±0.0082 | 0.10±0.0011 | 0.07±0.0091 | 0.24±0.077 | 1 |
| #7+Our Method | **0.12**±0.0063 | **0.23**±0.0030 | **0.22**±0.015 | **0.07**±0.0031 | **0.30**±0.061 | **0.85**±0.71 |

Table 11. Evaluation of $\tau$ on CASIA

| Method | JC | FMI | NMI | ARI | ACC |
|---|---|---|---|---|---|
| #2~#4 | 0.24 | 0.23 | 0.56 | 0.66 | 0.22 |
| Our Method | **0.07** | **0.02** | **0.17** | **0.38** | **0.12** |

\* $\tau$ is computed with all generated indexes of all random sets.

Table 10 shows the detailed clustering evaluation indexes. Table 11 shows the index difference rates with different features. Although CASIA supplies with more complicated face images, the results of the proposed method can be still achieved higher in external indexes and lower in internal indexes with less exceptions.

### 4.4.6    Performance on MNIST

Different from all the other datasets applied in sections 4.4.1~4.4.5, MNIST is a dataset of handwritten digit images. The images in MNIST are both simple and various. There are no related background information or background noise in each digit image. Thus MNIST is an ideal dataset for image classification. Meanwhile it is considered as a pre-processed dataset for image clustering, and every type of classical features can achieve good results including RP. For all images with a size of 28×28 pixels, the clustering based on DCN has proved to be well implementable.

Here we set $\Delta$ to 0.2 and $\delta$ to 0.1, and the cluster number is set to 10 for K-means and GMM.

The results of performance evaluation with the proposed method except DCN-based method are displayed in table 12. Table 13 shows the index difference rates with different features.

Table 12 shows, the results generated by transfer learning based clustering are improved. Although the initial features have effects on clustering, the proposed method can weaken the effects a lots. Of course, better initial features guarantee to achieve labels that are more correct and much better results.

DCN uses a forward network with 4 layers and sets the numbers of neurons to be 500, 500, 2,000, and 10. The reproducible codes remain mostly unchanged from https://github.com/boyangumn/DCN. However, we failed in reproducing clustering results based on #9. Therefore, we use their results directly and simulate correct labels according to its ACC performance in raw MNIST. Meanwhile when we change one cluster image with a size of 28×28 pixels to another one with a size of 256×256 pixels, DCN faces the curse of dimensionality. And the configurations for the depth of layers and number for each layer are unavailable. We use K-means instead of DCN in optimization part. It is pity that DCN can be comparable only in the experiments related to the image clustering based on MNIST.

Table 12 Clustering evaluation on raw MNIST

| Method | JC | FMI | NMI | ARI | ACC | DBI |
|---|---|---|---|---|---|---|
| #1 | 0.29 | 0.45 | 0.52 | 0.39 | 0.55 | 1.0 |
| #1+Our Method | **0.39** | **0.56** | **0.62** | **0.50** | **0.65** | **0.75** |
| #2 | 0.32 | 0.48 | 0.55 | 0.42 | 0.58 | 1.0 |
| #2+Our Method | **0.39** | **0.56** | **0.64** | **0.51** | **0.65** | **0.84** |
| #3 | 019 | 0.32 | 0.36 | 0.25 | 0.46 | 1.0 |
| #3+Our Method | **0.34** | **0.51** | **0.59** | **0.45** | **0.59** | **0.71** |
| #4 | 0.29 | 0.45 | 0.51 | 0.39 | 0.57 | 1.0 |
| #4+Our Method | **0.39** | **0.55** | **0.63** | **0.50** | **0.64** | **0.84** |
| #7 | 0.14 | 0.26 | 0.19 | 0.14 | 0.35 | 1 |
| #7+Our Method | **0.19** | **0.36** | **0.26** | **0.23** | **0.52** | **0.74** |

Table 13. Evaluation of $\tau$ on MNIST

| Method | JC | FMI | NMI | ARI | ACC |
|---|---|---|---|---|---|
| #1~#4 | 0.20 | 0.17 | 0.17 | 0.21 | 0.11 |
| Our Method | **0.06** | **0.05** | **0.04** | **0.06** | **0.05** |

Table 14 Comparative Results of the proposed method with DCN

| Index | #9 | Simulated #9 | Simulated #9+Our Method |
|---|---|---|---|
| JC | ✗ | 0.53±2.0e-04 | **0.75±0.020** |
| FMI | ✗ | 0.69±2.1e-04 | **0.86±0.010** |
| NMI | 0.81 | 0.66±1.9e-03 | **0.86±0.010** |
| ARI | 0.75 | 0.66±2.7e-04 | **0.84±0.010** |
| ACC | 0.83 | 0.83 | **0.92±0.010** |
| DBI | ✗ | 1.0 | **0.42±0.010** |

According to DCN's ACC index of raw MNIST, there are 83% correct labels in every cluster. Thus, we randomly choose 83% images from a same class in MNIST, and the rest of unchosen images are labeled as wrong labels. In this way, the initial evaluation indexes are worse than the actual results produced by DCN. Moreover, we repeat the simulation for more than three times to guarantee the effectiveness of the experiments.

The actual indexes, simulated indexes and indexes gained with the proposed method are shown in table 14. Although the simulated data are not perfectly reproduced, our method still improves the cluster performance.

### 4.5 Sensitivity of $\Delta$ and $\delta$

$\Delta$, theoretically represents the confidence values of clustering results. If the value of $\Delta$ is larger, then the initial clustering is supposed to perform more desirably. The progress can possess more time consumption without $\delta$. On the contrary, $\delta$ may lead to local results to be unsatisfying. Thus, it is necessary to examine the effect of these two parameters on the proposed method.

We just employ ACC as the model for performance evaluation in this part and perform the experiments with sampled Caltech 101 and Stanford Dogs. The results are shown in Figure 2.

$\delta$ can fasten the optimization with a relatively small number of iterations. Figure 3 shows counts of iterations with different $\delta$ and fixed $\Delta=0.2$.

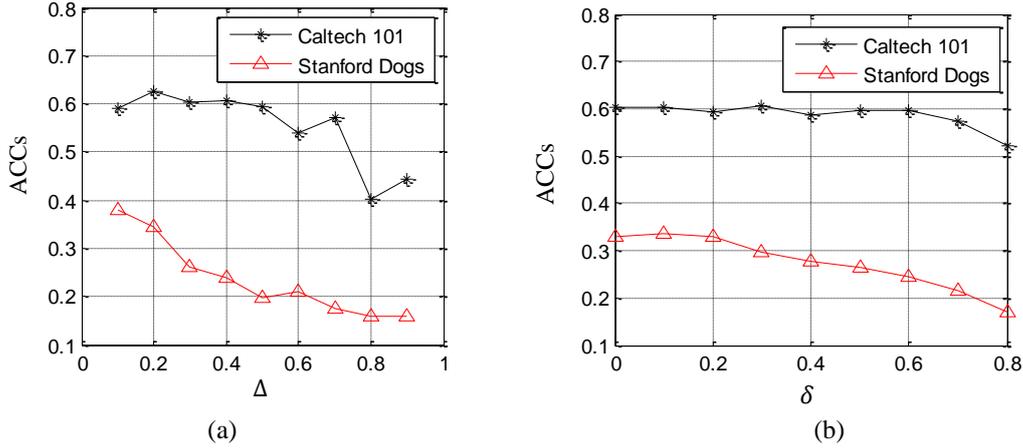

Fig. 2. ACCs of clustering Caltech 101 and Stanford Dogs, achieved respectively by (a) experiments with different $\Delta$ s and fixed $\delta = 0.1$, and (b) experiments with different $\delta$ s and fixed $\Delta = 0.2$

As shown in Figure 2, the configurations of $\Delta$ and $\delta$ have effects on the results. $\Delta$ is relatively more important to the result because a larger $\Delta$ can introduce more noisy labels in training. A proper $\Delta$ can ensure effectiveness of learned features. When little prior knowledge is supplied, it is a suitable choice to set $\Delta$ with values close to $1/K$, which is the ACC of a random clustering. A larger $\delta$ may make the proposed method stop faster, but accordingly $\delta$ is suggested to set with values close to 0.1. In this case, it can hardly affect the results and decrease the time consumption as Figure 2 shows.

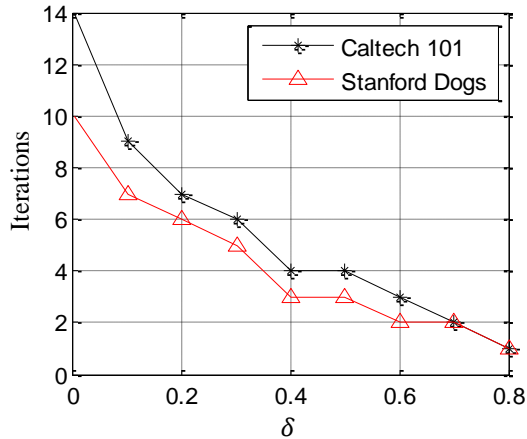

Fig. 3. Counts of iterations with different $\delta$ and fixed $\Delta = 0.2$ for experiments on Caltech 101 and Stanford Dogs.

### 4.6 Time consumption

In this section, we list the employed hardware equipment and software in table 15. Actually many factors can influence the time consumptions on implementing experiments, including the number of images in different datasets, images with different resolutions, and configuration parameters. In table 16, we show the maximum statistical time consumption on (usually goes with GMM) experiments with the employed datasets and the relative rough data.

The time consumptions are acceptable but not satisfying enough. Even though the number of images in FERET is set to 140, transfer learning by fine-tuning a pre-trained model still spends around half an hour. In addition, as shown in table 16, the implemented codes are combinations of matlab, python and C++ (Caffe), which means there are extra more time consumptions on data conversion. Maybe we will later try to find out a faster way on image clustering based on transfer learning.

Table 15 Configurations of the implemental experiments

| Name | Configures |
|---|---|
| CPU | Intel (R) Core(TM) i7-6700 |
| RAM | 32.0GB |
| Operating System | Windows 10 Professional |
| GPU | NVIDIA GeForce GTX 1080 Ti |

Table 16 Time consumptions of experimental in section 4.4

| Dataset | Time Consumption | Number of Images |
|---|---|---|
| MNIST | 455± 10 minutes | 70,000 |
| Caltech Categories | 79 ± 5 minutes | 3,648 |
| Caltech 101 | 38 ± 4 minutes | 1,126 |
| Stanford Dogs | 101± 7 minutes | 3,352 |
| FERET | 28± 3 minutes | 140 |
| CASIA-WebFace | 67± 5 minutes | 1,317 |

* Number of images refers to the employed sample images in experiments, which may be smaller than the actual number of original datasets.

## 4. Conclusion

The method of image clustering based on Transfer Learning is proposed in this paper. The proposed method can be validated by implementing a large number of experiments and employing different types of evaluation models that it can improve the original clustering results and weaken the effect of the initial features on image clustering. The method is also robust to the variations of different image categories. We believe that dynamic feature dimensions and faster implementations are highly demanded for image clustering in the future, which may be the next stage of studies in this field.